*Invited paper*

# Handshape recognition for Argentinian Sign Language using ProbSom

Franco Ronchetti, Facundo Quiroga, César Estrebou, and Laura Lanzarini

*Instituto de Investigación en Informática LIDI, Facultad de Informática,*
Universidad Nacional de La Plata
{fronchetti,fquiroga,cesarest,laural}@lidi.unlp.edu.ar

## Abstract

Automatic sign language recognition is an important topic within the areas of human-computer interaction and machine learning. On the one hand, it poses a complex challenge that requires the intervention of various knowledge areas, such as video processing, image processing, intelligent systems and linguistics. On the other hand, robust recognition of sign language could assist in the translation process and the integration of hearing-impaired people.

This paper offers two main contributions: first, the creation of a database of handshapes for the Argentinian Sign Language (LSA), which is a topic that has barely been discussed so far. Secondly, a technique for image processing, descriptor extraction and subsequent handshape classification using a supervised adaptation of self-organizing maps that is called ProbSom. This technique is compared to others in the state of the art, such as Support Vector Machines (SVM), Random Forests, and Neural Networks.

The database that was built contains 800 images with 16 LSA conjurations, and is a first step towards building a comprehensive database of Argentinian signs. The ProbSom-based neural classifier, using the proposed descriptor, achieved an accuracy rate above 90%.

**Keywords:** handshape recognition, sign language recognition, SOM, Radon transform

## 1 Introduction

Automatic sign recognition is a complex, multidisciplinary problem that has not been fully solved. Even though there has been some progress in gesture recognition, driven mainly by the development of new technologies, there is still a long road ahead before accurate and robust applications are developed that allow translating and interpreting the gestures performed by an interpreter[1]. The complex nature of signs draws effort in various research areas such as human-computer interaction, computer vision, movement analysis, automated learning and pattern recognition. Sign language, and the Argentinian Sign Language (LSA) in particular, is a topic that is currently being promoted by governments and universities to foster the inclusion of hearing impaired people in new environments. There is little documentation and even less information digitally available.

The full task of recognizing a sign language gesture involves a multi-step process [1]:

- Locating the hands of the interpreter
- Recognizing the shapes of the hands
- Tracking the hands to detect the movements performed
- Assigning a semantic meaning to the movements and shapes
- Translating the semantic meaning to the written language

These tasks can be carried out and assessed separately, since each has its own unique complexity. There are several approaches for tracking hand movements: some use 3D systems, such as MS Kinect, and others simply use a 2D image from an RGB camera. There are even systems with movement sensors such as special gloves, accelerometers, etc.

This paper focuses on the problem of classifying handshapes. In particular, it focuses on the extraction of characteristics that are representative of the hand and allow recognizing such handshapes using a variation of a supervised competitive neuronal network called ProbSom [2]. The objective of this work is to generate a processing sub-unit for the automated recognition of signs, such as those introduced by [3] sub-unit to subdivide sign recognition into modules.

Sign languages are different in each region of the world, and each has their own lexicon and group of signs. Thus, sign language recognition is a problem that needs to be tackled differently in each region, since new gestures or handshapes involve new challenges that were not considered before. In particular, there are almost no systems or databases that represent the group of gestures used in the Argentinian Sign Language (LSA). In this paper, we also tackle the development of a database with 16 LSA handshapes interpreted by 10 different individuals. The images obtained were then used for extracting characteristics and the subsequent classification process.

There are numerous publications dealing with the automated recognition of sign languages, and [1] presents a review of the state of the art in sign language recognition. However, the experiments in each article applies to a specific scenario that sometimes is not possible to replicate exactly, and when it can be replicated, there are some limitations. For instance, different articles use depth sensors, such as MS Kinect or other similar ones, to capture 3D images. [4],[5] and [6] use depth images to sort American Sign Language (ASL) handshapes. In general, these approaches have two problems: a machine with similar features as the one used for testing is required, and these devices still are affected by a high error rate to calculate depth images. Other approaches, such as the one presented here, use only RGB images. [7] creates a skin color probabilistic model to detect and track the hands of an interpreter on a video. [8] uses this model to segment the hands and apply a classifier based of Markov models. In general, the systems that are based only on skin color are not robust for background variations or interpreter clothes, as well as hand-hand or hand-face occlusions. Usually, they require the addition of morphological information to the color filtering to identify the position and shape of the hand.

This document is organized as follows: Section 2 describes the LSA database, and the image processing, hand feature extraction, and classification model used. Section 3 details the experiments carried out, and Section 4 presents the general conclusions.

## 2 Methods

### 2.1 Argentinian Sign Language Handshapes Database (LSA16)

The handshape database for the Argentinian Sign Language [1], created with the goal of producing a dictionary for LSA and training an automated sign translator, includes 800 images where 10 subjects executed 5 repetitions of 16 different types of handshapes used in the various signs in that language. Handshapes were selected among the most commonly used ones in the LSA lexicon, and they can be seen in Figure 1. Each handshape was executed repeatedly in various positions and rotations on the perpendicular plane to the camera to increase diversity and realism in the database.

---

[1] More information about this database can be found at http://facundoq.github.io/unlp/lsa16/.

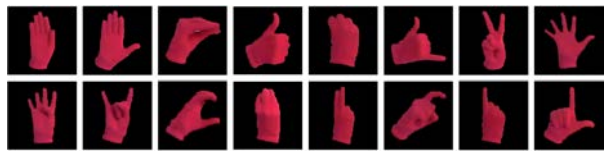

Figure 1: Examples of each class in the LSA16 database

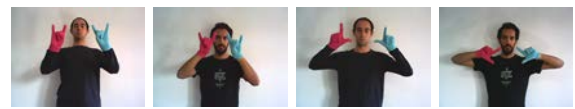

Figure 2: Unsegmented images in the LSA16 database

Subjects wore black clothes and executed the handshapes on a white background with controlled lighting, as shown in Figure 2. To simplify the problem of hand segmentation within an image, subjects wore fluorescent-colored gloves. The glove substantially simplifies the problem of recognizing the position of the hand, and removes all issues associated to skin color variations, while fully retaining the difficulty of recognizing the handshape.

### 2.2 Preprocessing and Descriptors

Below, we present the pre-processing activities carried out with the database's segmented hand images, as well as the descriptors calculated based on this images, and the classification model presented. The input to the pre-processing stage is an image where the only non-black pixels are those corresponding to the hand, as shown in figure 1.

**Pre-processing**

For each image, the signgle biggest connected component is determined to obtain the hand segmentation mask. Afterwards, the main axes of the pixels are calculated, which are used to determine hand inclination $\phi$.

The image is then rotated by $-\phi$ to put it in a canonical orientation. Since this orientation is not sensitive to 180° hand rotations, the image may be in an upwards or downwards position. To correct this, the number of possible crosses of each horizontal line is determined for the image, and the position of the fingers is estimated based on the mode of the number of crosses (ie, whether they are in the upper or lower part of the image).

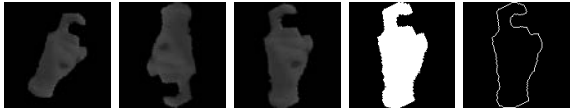

Figure 3: From left to right: Segmented image, oriented image, image with corrected rotation, hand segmentation mask, and contour.

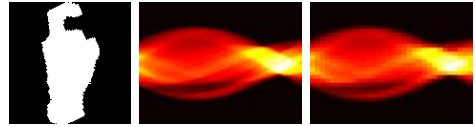

Figure 4: Original image, Radon Transform, resampled Radon Transform

The image is resampled without affecting its aspect ratio to $128 \times 128$ pixels and is centered. The contour of the hand is obtained by applying a border filter to the hand segmentation mask.

**Descriptors**

Below, two descriptors are described, one based on Radon Transform and one based on Scale-Invariant Feature Transform (SIFT).

**Radon Transform** The Radon Transform has been used in the past for recognizing objects as well as identifying individuals based on the characteristics of their hands [4].

The Radon Transform of a 2D image $f : \mathbb{R}^2 \to \mathbb{R}$ is defined as a line integral on the image. The line $L$ used for integration is given by a pair $(b, \theta)$, where $b$ is the distance to the origin of the line and $\theta$ is the angle with the horizontal axis of the image. It is given by the following equation:

$$R_{(b,\theta)} = \iint_{L_{(b,\theta)}} f(\vec{x})|d\vec{x}| = \int_{-\infty}^{\infty} f(x(t), y(t))dt$$
$$= \int_{-\infty}^{\infty} f(tsin\theta + bcos\theta, -tcos\theta + bsin\theta)dt$$

By applying the corresponding discrete version to the segmented image for all possible integer number combinations of $(b, \theta)$ (1..180 for $\phi$, a value $K$ dependent on the size of the image for $b$), a descriptor $R \in \mathbb{R}^{180 \times K}$ is obtained. Then, to reduce its dimensionality, $R$ is resampled to a constant size matrix $r \in \mathbb{R}^{32 \times 32}$. This descriptor can be used as global, considering it as a vector $r' \in \mathbb{R}^{32^2}$, or as 32 local descriptors by considering each row $r_i$, $i = 1, \ldots, 32$, $r_i \in \mathbb{R}^{32}$ as a descriptor. Each $r_i$ then contains a soft approximation to $R_{(b,\theta)}$ for all $b$, where $\phi$ corresponds approximately to the mean of a subset of adjacent angles.

In particular, since the classifier presented here, ProbSom, uses a set of arbitrary cardinality vectors as input, the $r_i$ vectors were used for it, and the full vector global matrix $r'$ was used for the rest of the classifiers that were tested.

**SIFT** A SIFT descriptor is a 3D spatial histogram of the gradients in an image that characterizes the aspect of a point of interest. To do so, the gradient of each pixel is used to calculate a more elementary descriptor formed by the location of the pixel and the orientation of the gradient. Given a potential point of interest, these elementary descriptors are weighed by the gradient norm and accumulated in a 3D histogram that represents the SIFT descriptor of the region around the point of interest. When building the histogram, a Gaussian weight function is applied to the elementary descriptors to reduce the significance of the gradients that are farther away from the center of the point of interest.

SIFT descriptors have been applied to various computer vision tasks, including handshapes recognition [10] and face recognition [11].

## 2.3 ProbSom Classification Model

ProbSOM [2] is a probabilistic adaptation of Kohonen's self-organizing maps (SOMs)[12]. These maps are competitive, unsupervised networks that configure their neurons to represent the distribution of the input data processed during the training phase. As a result of this learning phase, a network is obtained where each neuron learns to represent an area of the input space, grouping data vectors by similarity or proximity.

ProbSOM's training process is carried out in the same way as that for the traditional SOM algorithm. ProbSOM has an additional stage after training to weight each neuron's representation ratio. To do so, all input patterns are re-processed and information is added to each winning neuron to indicate the class it represents and in which proportion.

The recognition process is also similar to that in SOM. The response mechanism that identifies the classes is a probabilistic system. Since each vector in itself does not allow identifying the classes, a sequence of vectors is required. When a set of characteristic vectors are entered into the network, a set of winning neurons is obtained, where each neuron represents several classes in a given ratio. The class is identified as that whose aggregated ratio is the maximum.

ProbSOM has proven to be a robust algorithm to solve classification problems [11, 2, 13] where samples are represented by a set of characteristic vectors, where the set has an arbitrary cardinality which depends solely on the sample.

Table 1: Random cross-validation accuracy for the LSA16 database.

| Method | CV performance |
| --- | --- |
| ProbSom with Radon | 92.3($\pm$2.05) |
| ProbSom with SIFT | 88.7($\pm$2.50) |
| Random Forest with Radon | 91.0($\pm$1.91) |
| SVM with Radon | 91.2($\pm$1.69) |
| Neural Net with Radon | 78.8($\pm$3.80) |

## 3 Results

### 3.1 Methodology and Results

In this section, we compare the performance obtained with the tests carried out with various methods and descriptors. In the case of ProbSom, tests were run with SIFT and Radon descriptors. Also, in the case of the descriptor based on Radon, tests were carried out using the standard models of the state of the art – Support Vector Machines (SVM), Random Forest, and Feedforward Neural Networks [2]. In the case of the Radon-only tests with other models, we tuned hyper-parameters and report only the best results obtained.

Table 1 shows the average recognition accuracy obtained under stratified random cross validation with n = 30 separate repetitions, using 90% of the images to train and 10% of the images for evaluation. These results show that ProbSOM has a performance that is comparable to that of other classification techniques. On the other hand, Radon descriptors proved to be much more representative than SIFT vectors. This could be due to the fact that, in general, SIFT descriptors are employed to identify interest points that are unique and then match two images if the interest points are similar in both images. The images in LSA16 may have several interest points (such as the tips of the fingers) that are common to many classes, which hinders the use of SIFT as it was used in [11] to recognized faces, using the same classification model.

**Inter-Subject Cross-validation** Using the best configuration obtained (Radon and ProbSOM descriptor), a cross-validation, inter-subject experiment was carried out, leaving one subject for testing and training the others. The mean for the 10 subjects with $n = 30$ separate repetitions was 87.9% ($\pm$4.7%). As expected, by leaving out one subject, the accuracy decreases, since each individual executes handshapes in their own specific way, with hand sizes and appearance changing from one individual to another. However, the system still yielded good results, and was able recognize a handshape executed by a new individual unknown to the system with good accuracy. Figure 5 shows the results obtained for each individual in the database.

---
[2] Additionally, tests were carried out using Fourier, Gabor Filter Bank and Local Binary Patterns descriptors (not described in this paper), but their results were worse than those presented here in almost all of the cases.

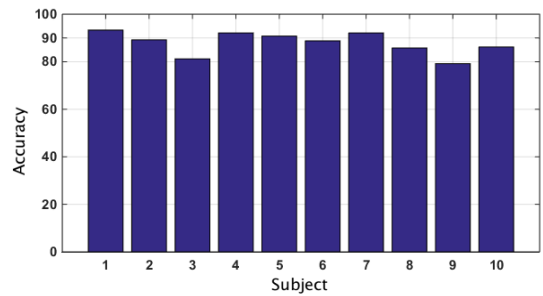

Figure 5: Inter-subject cross validation accuracies for LSA16.

### 3.2 Discussion

The descriptors used, together with the classification model, proved to be robust for classifying the handshapes in LSA16, even with inter-subject validation, which means that new individuals unknown to the system can be added. It should be noted that the success rate is similar for all classes in the database.

Since ProbSOM works probabilistically by building a ranking of possible candidate classes, it would be interesting to see what happens with the images that were miss-classified by the system. In those cases, if given an example we process it with ProbSOM and consider a success guessing with the best or second-best accuracy class output by the model, the accuracy goes up from 92.25% to 96.60%. This is very interesting if the model works as a dictionary, since the probability of the model could be used to show one or two (or more) possibilities. Similarly, a boosting scheme could be applied using a more specific classifier afterwards to solve any ambiguities if necessary.

## 4 Conclusion

We have presented a handshape database for the Argentinian Sign Language (LSA), as well as a model to pre-process handshape images and classify them.

The results obtained with the classification experiments were favorable, showing high accuracy both for random as well as inter-subject validation. Comparisons were also made with various existing descriptors and classification methods.

The model presented can be used to generate a lexical sub-unit as part of a general descriptor for LSA signs. In the future, we are planning to test the technique with other existing databases to determine is applicability, as well as extending it to use depth sensor images.